\documentclass[letterpaper]{article} 
\usepackage{aaai24}  
\usepackage{times}  
\usepackage{helvet}  
\usepackage{courier}  
\usepackage[hyphens]{url}  
\usepackage{graphicx} 
\usepackage{natbib}  
\usepackage{caption} 
\usepackage{algorithm}
\usepackage{algorithmic}

\usepackage{lipsum}

\usepackage{times}
\usepackage{epsfig}
\usepackage{graphicx}
\usepackage{amsmath}
\usepackage{amssymb}
\usepackage{makecell}
\usepackage{amsmath,amssymb,stmaryrd}
\usepackage{booktabs}
\usepackage{multirow}
\usepackage{graphicx}
\usepackage{subfigure}
\usepackage{amsfonts}
\usepackage{arydshln}

\usepackage{newfloat}
\usepackage{listings}
\DeclareCaptionStyle{ruled}{labelfont=normalfont,labelsep=colon,strut=off} 
\floatstyle{ruled}
\newfloat{listing}{tb}{lst}{}
\floatname{listing}{Listing}
\title{Improving Audio-Visual Segmentation with Bidirectional Generation}
\author{
    \textsuperscript{\rm 2}Dawei Hao\equalcontrib,
    \textsuperscript{\rm 1,3}Yuxin Mao\equalcontrib,
    \textsuperscript{\rm 4}Bowen He,
    \textsuperscript{\rm 1}Xiaodong Han,
    \textsuperscript{\rm 3}Yuchao Dai,
    \textsuperscript{\rm 1}Yiran Zhong\thanks{Corresponding author. Email: \texttt{zhongyiran@gmail.com}.}
}
\affiliations{
    \textsuperscript{\rm 1}OpenNLPLab, Shanghai AI Lab, 
    \textsuperscript{\rm 2}Bilibili Inc., 
    \textsuperscript{\rm 3}Northwestern Polytechnical University, 
    \textsuperscript{\rm 4}NIO    

}

\usepackage{bibentry}

\begin{document}

\maketitle

\begin{abstract} 
The aim of audio-visual segmentation (AVS) is to precisely differentiate audible objects within videos down to the pixel level. Traditional approaches often tackle this challenge by combining information from various modalities, where the contribution of each modality is implicitly or explicitly modeled. Nevertheless, the interconnections between different modalities tend to be overlooked in audio-visual modeling. In this paper, inspired by the human ability to mentally simulate the sound of an object and its visual appearance, we introduce a bidirectional generation framework. This framework establishes robust correlations between an object's visual characteristics and its associated sound, thereby enhancing the performance of AVS. To achieve this, we employ a visual-to-audio projection component that reconstructs audio features from object segmentation masks and minimizes reconstruction errors. Moreover, recognizing that many sounds are linked to object movements, we introduce an implicit volumetric motion estimation module to handle temporal dynamics that may be challenging to capture using conventional optical flow methods. To showcase the effectiveness of our approach, we conduct comprehensive experiments and analyses on the widely recognized AVSBench benchmark. As a result, we establish a new state-of-the-art performance level in the AVS benchmark, particularly excelling in the challenging MS3 subset which involves segmenting multiple sound sources. Code is released in: \url{https://github.com/OpenNLPLab/AVS-bidirectional}.
\end{abstract}

\section{Introduction}
The foundation of human perception heavily relies on sight and hearing, which together absorb a substantial amount of external information. Integrating audio and visual information in a collaborative manner plays a crucial role in enhancing human scene understanding capabilities. Our daily experiences demonstrate that both auditory and visual cues contribute to our understanding of the concepts of objects. Therefore, audio-visual learning is essential in enabling machines to perceive the world through multi-modal information as humans do.

In the realm of audio-visual learning, the quest to disentangle auditory entities within videos at the minutiae of pixel level has given rise to the field of audio-visual segmentation (AVS). The pursuit of this goal has spurred the development of a myriad of techniques, often grounded in the fusion of multi-modal features, where the contribution of each modality is implicitly or explicitly modeled. However, a notable weakness in current methodologies is the insufficient attention given to the intricate relationship between various modes in audio-visual modeling. This lacuna forms the focal point of this paper.

Drawing inspiration from the remarkable human ability to conjure auditory perceptions corresponding to visual stimuli and vice versa, our study delves into a novel paradigm. However, directly replicating this procedure in a network,~\emph{i.e.,} generating audio from object segmentation masks, is difficult because the network needs to have audio creation capability, which requires a substantial amount of data to earn.  
Instead, we propose a bidirectional generation schema in feature space, meticulously designed to forge robust correlations between the visual attributes of objects and their corresponding auditory manifestations. 
Specifically, we utilize a visual-to-audio projection module to reconstruct audio features from object segmentation masks and minimize reconstruction errors. This schema allows the model to build a strong correlation between visual and audio signals. Furthermore, we recognize the profound relationship between sound and motion and introduce an implicit volumetric motion estimation module to address motions that may be difficult to capture using optical flow approaches~\cite{zhong2019unsupervised,wang2020displacement,zhong2022displacement}. We construct a visual correlation pyramid for input video frames, make use of the inter-frame motion information to smooth the significant angle motion, and get clearer masks for dynamic objects.             

To substantiate the efficacy of our novel framework, we undertake an extensive series of experiments and rigorous analyses on the AVSBench benchmark~\cite{zhou2022audio}. Our efforts culminate in the establishment of unprecedented benchmark performance, particularly within the challenging MS3 subset. In the spirit of scientific transparency and reproducibility, we commit to releasing both the code and the pre-trained models in the near future. 

The key contributions of our work include:
\begin{itemize}
    \item Proposed an innovative, efficient audio-visual segmentation approach using bidirectional generation supervision, which builds strong correlations between audio-visual modalities. 
    \item Constructed a visual correlation pyramid for input video frames, leveraging implicit inter-frame motion information to enhance object mask quality.
    \item Achieved state-of-the-art performance on the AVSBench benchmark.
\end{itemize}

\begin{figure*}[t]
\begin{center}
\centering
  \includegraphics[width=0.85\linewidth]{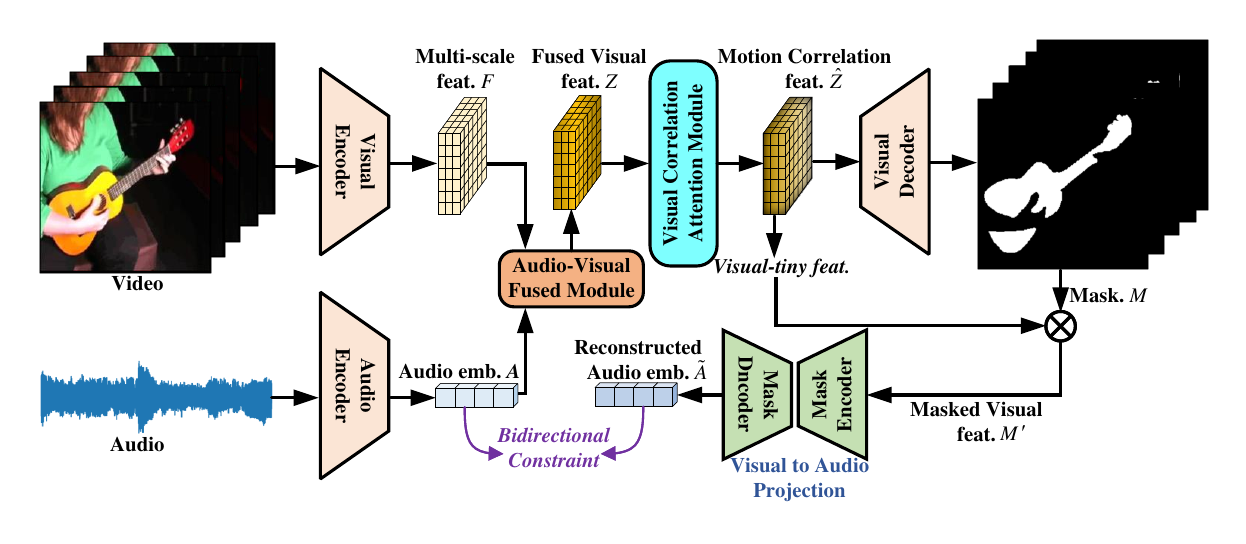}
  \vspace{-8mm}
  \caption{Overview of the proposed model, which follows a hierarchical Encoder-Decoder pipeline. The encoder inputs the video frames and the entire audio clip, and outputs multi-scale visual and audio features, respectively denoted as ${F}$ and ${A}$. The multi-scale visual feature ${F}$ and audio embedding ${A}$ are further sent to the audio-visual fused module, which builds the audio-visual mapping to assist with identifying the sounding object. The visual correlation attention module predicts inter-frame movement information resulting in motion correlation features $\hat{Z}$. The decoder progressively enlarges the fused feature maps and finally generates the output mask $M$ for sounding objects. In addition, masked visual features $M^{\prime}$
  is supplied to visual-to-audio projection module, in which the result reconstructed audio embedding $\widetilde{A}$ and the original audio embedding $A$ construct bidirectional constraint together.}
  \vspace{-8mm}
  \label{fig:2}
  \end{center}
\end{figure*}

\section{Related Work}
\textbf{\textit{Audio-Viusal Segmentation.}}~More challenging than the sound source localization~(SSL)~\cite{chen2021localizing,cheng2020look,hu2020discriminative,qian2020multiple}, event parsing~(AVP)~\cite{wu2021exploring, tian2020unified,zhou2023improving} and event localization (AVEL)~\cite{zhou2021positive,zhou2022contrastive} task is audio-visual segmentation task. 
These methods require the fusion of audio and visual signals. Such as audio-visual similarity modeling by computing the correlation matrix~\cite{arandjelovic2017look,arandjelovic2018objects}, audio-visual cross attention~\cite{xuan2020cross}, audio-guided Grad-CAM~\cite{qian2020multiple}, or using a multi-modal transformer for modeling the long-range dependencies between elements across modalities directly. Zhou et al.~\cite{zhou2022audio} released a segmentation dataset with pixel-level annotation of localization, and they introduced audio-visual segmentation (AVS). The audio-visual segmentation~\cite{mao2023multimodal,mao2023contrastive} requires locating the actual sounding object(s) from multiple candidates and describing the contours of multiple sounding objects clearly and accurately. 

\noindent
\textbf{\textit{Motion Estimation.}}~To model motion information between video frames, a more traditional method is to explicitly use the method based on optical flow to model the pixel-by-pixel intensive correspondence between adjacent video frames \cite{tokmakov2017learning,yang2021self,wang2020displacement}. However, for fast-moving objects and dynamic video scenes with occluded objects, the error of optical flow estimation accumulates, leading to the wrong analysis of motion information between two frames.

\noindent
\textbf{\textit{Bidirectional Consistency.}}~In addition to modeling inter-frame motion estimates, our framework also incorporates cycle consistency as a supervisory signal. \citet{hu2022mix} formulated the image and sound as graphs and adopted a cycle-consistent random walk strategy for separating and localizing mixed sounds. 
This method only locates the sounding area in videos but does not consider the shape of objects. 
In this paper, we introduce audio-visual-audio loop consistency constraints to build strong correlations between audio and segmented masks and strengthen audio supervision throughout the prediction process.

\section{Method}
In this section, we give a thorough explanation of the proposed method by first giving a general overview of the model's structure and then going into detail about each of its constituent parts. 

\subsection{The overall architecture}
We illustrate the overall architecture of our proposed model in Figure~\ref{fig:2}. 
It consists of the following major parts: (1) Visual Encoder, (2) Audio Encoder, (3) Audio-Visual Fused Module, (4) Visual Correlation Attention Module, (5) Visual to Audio Projection Module, and (6) Visual Decoder.

\noindent
\emph{\textbf{Information Flow.}}
The process begins with the encoder receiving video frames and the entire audio clip. It then outputs multi-scale visual and audio features, represented as ${F}$ and ${A}$, respectively. These features are then sent to the audio-visual fused module, which creates an audio-visual map to aid in identifying the sounding object. The visual correlation attention module predicts inter-frame movement information to produce motion correlation features $\hat{Z}$. The decoder then progressively enlarges the fused feature maps to generate the output mask $M$ for sounding objects. Additionally, masked visual features $M^{\prime}$ are provided to the visual-to-audio projection module, which uses them to reconstruct the audio embedding $\widetilde{A}$, while also considering the original audio embedding $A$ to create a bidirectional constraint.

We define a video clip with a length of ${T}$ as ${{\left\{ \left\{I_t\right\},\left\{A_t\right\}\right\}}_{t=1}^{T}}$, and the video frame ${I_t}\in \mathbb{R}^{3 \times H \times W}$ corresponding to the audio ${A_t}$ at the moment ${t}$ as the reference frame, ${\left\{A_t\right\}_{t=1}^{T}}$ represents the audio with length $T$, and the pixel-level mask ${\left\{M_t\in\left\{0, 1\right\}\right\}_{t=1}^{T}}$ corresponding to the network as the output. ${H,W}$ represent the height and width of the frame, respectively. 

\subsection{Visual Encoder} 
We use Pyramid Vision Transformer (PVT-v2)~\cite{wang2022pvt} as our visual encoder. We load the pre-trained weights of ImageNet~\cite{russakovsky2015imagenet} to speed up network convergence and improve performance. The input of our visual encoder is the $T$ video frames ${I\in \mathbb{R}^{T \times 3 \times H \times W}}$. The output is visual features of four scales ${F_i}\in{\mathbb{R}^{T \times {C_i} \times {h_i} \times {w_i}}}$, where ${{\left({h_i}, {w_i}\right)}=\left(H,W\right) \big /{2^{i+1}}}$, ${i=1,2,3,4}$. ${C_i}$ is feature channel and ${T}$ represents the length of the video.

\subsection{Audio Encoder} We first process ${\left\{A_t\right\}_{t=1}^{T}}$ to a Mel-Spectrogram via the short-time Fourier, and then fed it into VGGish~\cite{hershey2017cnn} that is pre-trained on AudioSet~\cite{gemmeke2017audio} to obtain audio features ${A\in \mathbb{R}^{T \times d}}$, ${d=128}$ is the feature dimension.
\begin{figure}
\begin{center}
\centering
  \includegraphics[width=1\linewidth]{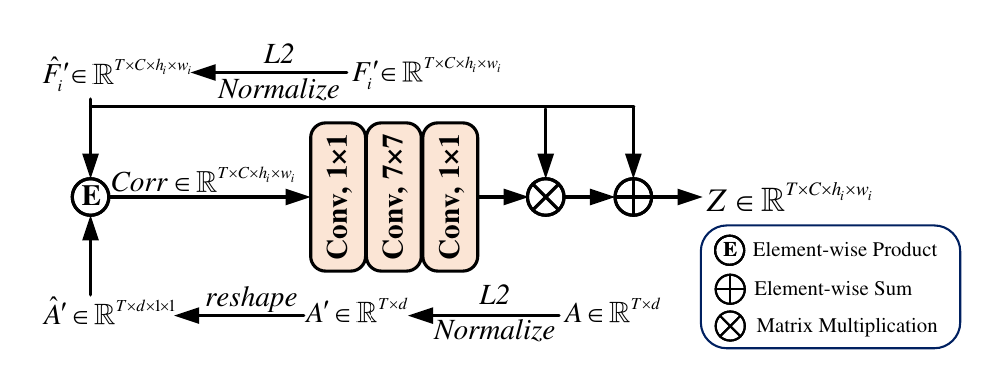}
  \vspace{-10mm}
  \caption{Audio-visual fused module takes the multi-scale features ${F}^{\prime}_{i}$ and audio features ${A}$ as inputs. The symbols "$\oplus$" and "$\otimes$" denote matrix multiplication and element-wise addition, respectively.}
  \vspace{-6mm}
  \label{fig:3}
  \end{center}
\end{figure}

\subsection{Audio-Visual Fused Module}
\label{Audio-Visual Fused Module}
We introduce a cross-modal fusion module for combining audio and video data, leveraging audio features to enhance vocal object prediction within video frames, as illustrated in Figure~\ref{fig:3}. Our approach involves several steps. Prior to fusion, we align the channel dimensions of multi-scale features extracted by the video encoder with those of both visual and audio features. This alignment yields transformed video features denoted as ${{F}^{\prime}_{i}\in \mathbb{R}^{T \times C \times {h_i} \times {w_i}}}$, where ${C=128}$ in our experimental setup. We then apply regularization and dimension transformation independently to the multi-scale video features ${{F_i}^{\prime}}$ and audio features ${A}$, leading to ${\hat{{F}^{\prime}_{i}}\in \mathbb{R}^{T \times C \times {h_i} \times {w_i}}}$ and ${\hat{{A}^{\prime}}\in \mathbb{R}^{T \times {d} \times {1} \times {1}}}$, respectively.

Utilizing the Enisum method, we perform element-wise multiplication and matrix operations between ${\hat{{F_i}^{\prime}}}$ and ${\hat{{A}^{\prime}}}$ across the spatial dimensions ${\left(h,w\right)}$, resulting in a correlation pyramid that captures the interplay between video and audio cues. This process culminates in the generation of fused video features denoted as ${{Z_i}\in \mathbb{R}^{T \times {C} \times {h_i} \times {w_i}}}$, as depicted below.

\begin{align}
\small
    \textrm{Corr} &= \displaystyle\sum_{h}\sum_{w}{{\hat{F}}_{i}^{\prime} \hat{{A}^{\prime}}}
\end{align}

\begin{figure}
\begin{center}
\centering
  \includegraphics[width=1\linewidth]{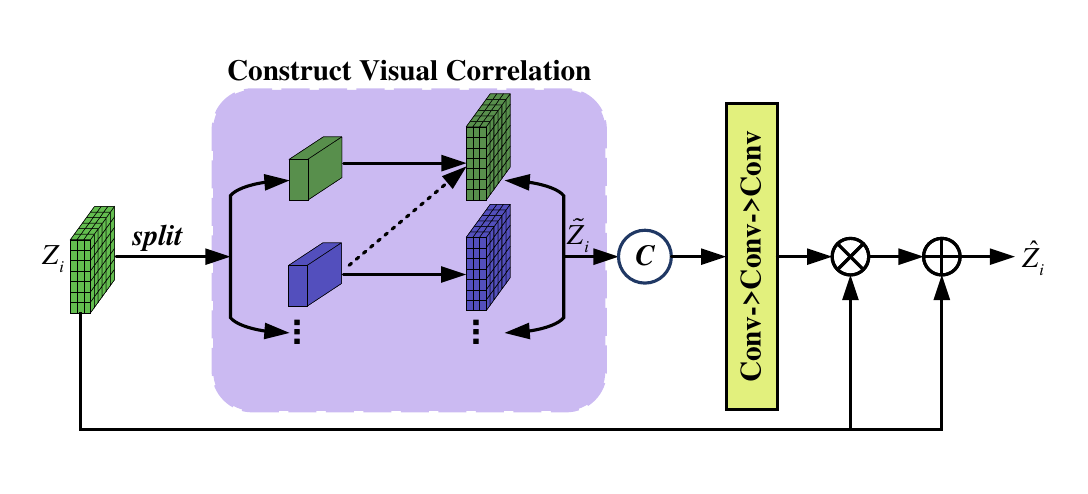}
  \vspace{-8mm}
  \caption{Visual correlation attention module integrates the video features ${Z_i}$ after audio-visual fusion, split the visual features ${Z_i}$ and construct the correlation pyramid ${\widetilde{Z}}_{i}$ of two adjacent frames.}
  \vspace{-6mm}
  \label{fig:4}
  \end{center}
\end{figure}

\subsection{Visual Correlation Attention Module}
\label{Visual Correlation Attention Module}
To improve the accuracy of our predictions and reduce any disruptions caused by significant movements between adjacent frames, we must consider the motion information between these structures. Our solution, as depicted in Figure~\ref{fig:4}, is an implicit correlation pyramid that can model motion and predict segmentation simultaneously. This network can optimize both motions and segmentation with only segmentation supervision. However, using optical flow methods would require ground truth optical flow to supervise the optical flow module. Otherwise, it may lead to errors accumulating during the continuous prediction process, which could adversely affect segmentation performance.

By integrating the video features ${{Z_i}\in \mathbb{R}^{T \times {C} \times {h_i} \times {w_i}}}$ of four different scales after cross-modal fusion, we split the visual features ${{Z_i}^{t}\in \mathbb{R}^{{C} \times {h_i} \times {w_i}}\left(t=1,\ldots, T\right)}$ of the moment ${t}$ according to the ${T}$ dimension, and then construct the correlation pyramid ${{\widetilde{Z}}_{i}^{t}\in \mathbb{R}^{{h_i} \times  {w_i} \times {h_i} \times {w_i}}}$ of two adjacent frames ${\left\{{Z_i}^{t}, {Z_i}^{t+1}\right\}_{t=1}^{T}}$. In the experiment, we constructed two cost calculations for the last two frames, and the calculation process is as follows:
\begin{equation}
\small
\begin{aligned}
    \{{\widetilde{Z}}_{i}^{t}\}_{t=1}^{T} &= 
    \{\textrm{Corr}({Z_i}^{t}, {Z_i}^{t+1})\in{\mathbb{R}}^{{h_i} \times  {w_i} \times {h_i} \times {w_i} }\}_{t=1}^{T} 
\end{aligned}
\end{equation}

By calculating the cost attention of visual features, we can obtain the motion correlation features ${\left\{\hat{Z}_{i}^{t}\in\mathbb{R}^{T \times {C} \times {h_i} \times {w_i}}\right\}_{t=1}^{T}}$. 

\begin{figure}
\begin{center}
\centering
  \includegraphics[width=1\linewidth]{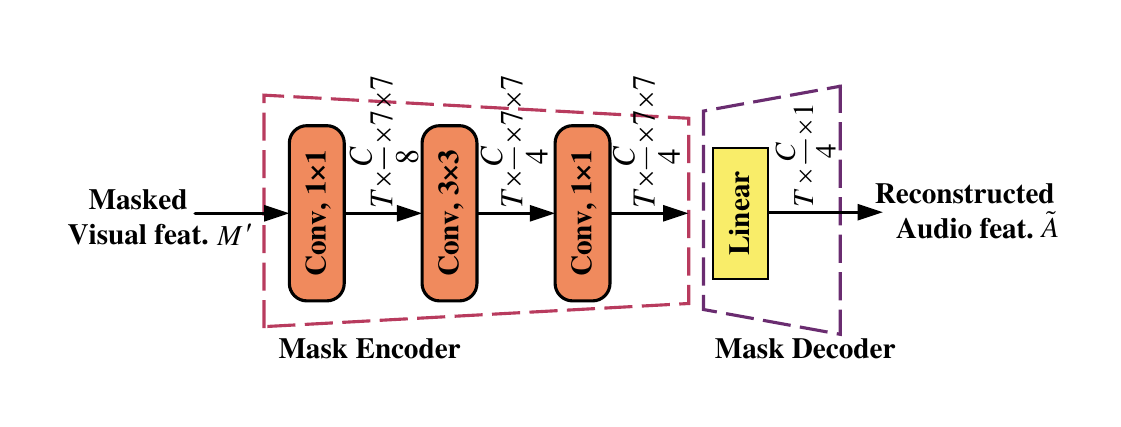}
  \vspace{-10mm}
  \caption{Visual to Audio Projection leverages masked visual features $M^{\prime}$
  to reconstruct audio embedding $\widetilde{A}$ and the original audio embedding $A$ construct bidirectional constraint together.}
  \vspace{-6mm}
  \label{fig:5}
  \end{center}
\end{figure}

\subsection{Visual to Audio Projection}
\label{Visual-to-Audio Projection}
To build correlations between object appearance and its audio, we propose a visual to audio projection model to reconstruct corresponding audio features according to the masked feature images, as shown in Figure~\ref{fig:5}.

We use the visual features $\hat{Z}_{i}$ obtained from the visual relevance attention module and select the minimum scale $\hat{Z}_{4}\in\mathbb{R}^{T \times \frac{C}{4} \times {h_4} \times {w_4}}$ as a part of the reconstructed object, where ${h_4}={w_4}=7$. We obtain the masked feature map ${M^{\prime}\in\mathbb{R}^{T \times \frac{C}{4} \times {7} \times {7}}}$ by calculating the correlation between the masked ground truth of actual sounding objects and the feature maps predicted by the segmentation network. We then input the reconstructed visual features $M^{\prime}$ into the mask Encoder and obtain ${\hat{M}}^{\prime}\in\mathbb{R}^{T \times \frac{C}{4} \times {7} \times {7}}$. Finally, we input ${\hat{M}}^{\prime}$ to the mask Decoder, which outputs the reconstructed audio features ${\widetilde{A}\in\mathbb{R}^{T \times {d} \times 1 \times 1}}$. In our experiment, the Single Source subset serves as the predictive mask, while for the multi-source subset, $M$ is the true mask.

\begin{figure*}[t]
\begin{center}
\centering
  \includegraphics[width=1\linewidth]{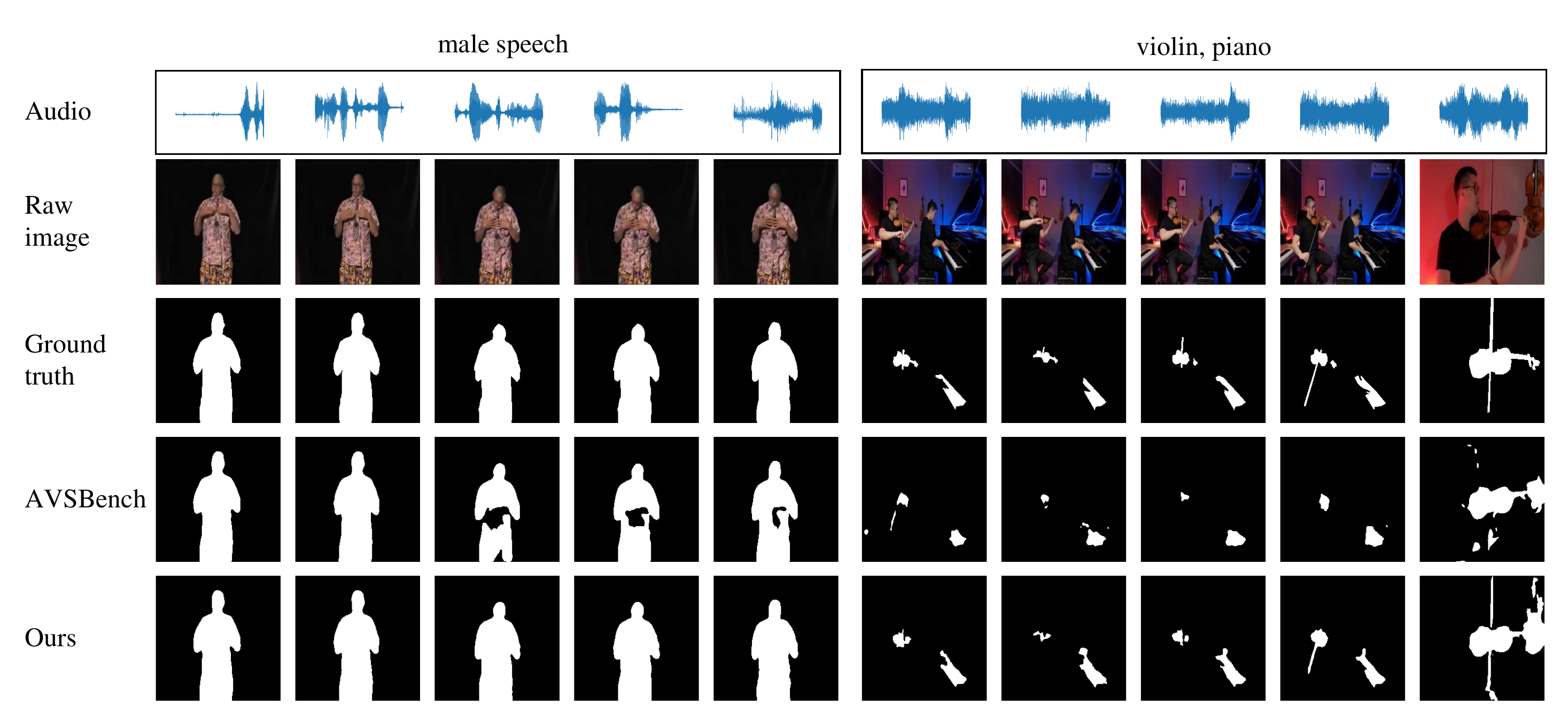}
  \vspace{-6mm}
  \caption{Qualitative examples of the AVSBench and our AVS framework. On the left is the output of the S4 setting, and on the right is the MS3 setting. AVSBench method only produces segmented maps that are not very precise, whereas our AVS framework can accurately segment the pixels of objects and create clear outlines of their shapes.}
  \vspace{-2mm}
  \label{fig:6}
  \end{center}
\end{figure*}

\subsection{Decoder}
In order to predict the sound object mask in a video frame, we utilize the Decoder structure explained in~\cite{zhou2022audio}. The decoded features are subsequently upsampled to the following stage, and the final result of the decoder is ${M \in {\mathbb{R}^{T \times {H} \times {W}}}}$.

\subsection{Loss objective} We adopt Binary cross entropy loss~(BCE) and Kullback-Leibler~(KL) divergence in~\cite{zhou2022audio} as part of the objective function, as shown in the formula, where ${avg}$ represents average pool operation and ${\otimes}$ is multiply operation element-by-element. The setting of hyper-parameter ${\lambda}$ is similar to that of the paper~\cite{zhou2022audio}. During the training of S4, ${\lambda=0}$. During the training in MS3, ${\lambda=0.5}$.
\begin{equation}
\small
\begin{aligned}
    \mathcal{L}_\text{AVS} &= \text{BCE}\left(M,Y\right) + \lambda\displaystyle\sum_{i=0}^n \left\{\text{KL}[avg(M \otimes Z), A]\right\}
 \end{aligned}
\end{equation}

To ensure the consistency between the reconstructed audio features ${\widetilde{A}}$ and the original input audio ${A}$ in the feature space, we designed latent loss ${L_\text{consistency}}$ to carry out consistency constraints on ${\widetilde{A}}$ and ${A}$, as shown in the Equ.~\ref{consistency_loss}.
\begin{equation}
\small
\begin{aligned}
    \mathcal{L}_\text{consistency} = \eta \textrm{Distance}\left[\textrm{Norm}\bf(A), \textrm{Norm}\bf(\widetilde{A})\right]
\label{consistency_loss}
 \end{aligned}
\end{equation}

\begin{table}[t]
\small
\centering
\caption{Impact of the different hyperparameter $\eta$ settings. }
\label{tab: table3}
\vspace{-2mm}
 \setlength{\tabcolsep}{3.8mm}
\begin{tabular}{cccccc}
\toprule
    \multirow{2}{*}{Settings} &\multicolumn{2}{c}{S4} &\multicolumn{2}{c}{\makecell[c]{MS3 \\From Scratch}} \\
    \cmidrule(r){2-3} \cmidrule(r){4-5}
     &mIoU &F-score &mIoU &F-score \\
        \midrule
            $\eta=0.1$ &81.40 &0.900 &53.07 &0.648 \\
            $\eta=0.5$ &81.61 &0.901 &54.35 &0.661 \\
            $\eta=1.0$ &\textbf{81.71} &\textbf{0.904} &\textbf{55.10} &\textbf{0.668}  \\
            $\eta=1.5$ &81.23 &0.901 &54.18 &0.665 \\            
\bottomrule
\end{tabular}
\vspace{-4mm}
\end{table}
In this paper, the expression calculates the norm along the last dimension, using the norm to calculate the regularized audio features $\widetilde{A}$ and $A$. We explored the influence of different hyperparameters $\eta$ on the performance of the model. According to the experimental results in Table~\ref{tab: table3}, we set the hyperparameter $\eta$ equal to $1.0$.
In summary, the total loss functions can be written as follows:
\begin{equation}
\small
\begin{aligned}
    \mathcal{L}_\text{AVS-BG} = \mathcal{L}_\text{AVS} + \mathcal{L}_\text{consistency}
 \end{aligned}
\end{equation}
\section{Experiments Results}
\subsection{Implementation details}
\paragraph{Dataset.} We conduct training and evaluation experiments on the AVSBench~\cite{zhou2022audio} dataset. AVSBench contains two subsets, namely Single-source~(S4) and Multi-sources~(MS3), depending on the number of sounding objects(s). 
The single-source subset contains $4932$ videos over $23$ categories, covering sounds from humans, animals, vehicles, and musical instruments. The multi-source subset contains $424$ videos and each video has multiple sounding sources and the sounding objects are visible in the frames. Each video was trimmed to $5$ seconds. 
There are two settings of audio-visual segmentation: 1) semi-supervised Single Sound Source Segmentation (S4), and 2) fully supervised Multiple Sound Source Segmentation (MS3). 
For the Single-source set, only part of the ground truth is given during training~(i.e., the first sampled frame of the videos) but all the video frames require a prediction during evaluation. In the Multi-sources subset, the labels of all five sampled frames of each are available for training. 
The goal for both settings is to correctly segment the sounding object(s) for each video clip by utilizing audio and visual cues.

\paragraph{Evaluation metrics.} We use the Mean Intersection-over-Union (mIoU) and F-score as the evaluation metrics. The former measures the contour accuracy of the predicted segmentation and the ground truth mask, and the latter considers both precision and recall. The F-score formulation is as follows, where ${\beta}^2$ is set to $0.3$ in our experiments.
\begin{equation}
\begin{aligned}
    F_{\beta} = \frac{\left(1+{\beta}^2\right)\times \text{precision}\times \text{recall}}{{\beta}^2 \times \text{precision+recall}}
 \end{aligned}
\end{equation}

\paragraph{Training Details.} 
We train our model using PyTorch on an NVIDIA Tesla V100 and utilize the Adam optimizer with a learning rate of ${10^{-4}}$.
The batch size is set to $8$, and we train on the Single-source subset for $15$ epochs and the Multi-sources subset for $30$ epochs. 
We resize all video frames to $224\times224$. 
Our experimental results are presented as percentages of the original results, and we use the PVT backbone to train all variations.

\subsection{Comparison with methods from related tasks}
\paragraph{Quantitative comparisons.} Following the comparison settings of AVSBench~\cite{zhou2022audio}, we compare the performance of our method with the AVSBench baseline model and state-of-the-art methods from other related tasks, shown in Table~\ref{tab: table1}. 
Our method consistently achieves significantly superior segmentation performance than the state-of-the-art methods, especially with $2.97$ higher mIOU under the Single-source set than the previous AVSBench method. Our approach surpasses SSL, VOS, and SOD methods by a wide margin, indicating that the inclusion of audio data boosts segmentation accuracy.
\begin{table}[h]
\small
\centering
\caption{Comparison with methods from related tasks. Results of mIoU and F-score under both S4 and MS3 are reported. We compare our method with LVS~\cite{chen2021localizing}, MSSL~\cite{qian2020multiple}, 3DC~\cite{mahadevan2020making}, SST~\cite{duke2021sstvos}, iGAN~\cite{mao2021transformer}, LGVT~\cite{zhang2021learning}, and AVS~\cite{zhou2022audio}.}
\label{tab: table1}
 \setlength{\tabcolsep}{1.5mm}
 \vspace{-2mm}
\begin{tabular}{llccccc}
\toprule
    \multirow{2}{*}{Tasks} & \multirow{2}{*}{Methods} & \multicolumn{2}{c}{S4} & \multicolumn{2}{c}{MS3} \\
    \cmidrule(r){3-4} \cmidrule(r){5-6}
    & & mIoU & F-score & mIoU & F-score \\
    \midrule
    \multirow{2}{*}{SSL} & LVS & 37.94 & 0.510 & 29.45 & 0.330 \\
                         & MSSL & 44.89 & 0.663 & 26.13 & 0.363 \\\hline
    \multirow{2}{*}{VOS} & 3DC & 57.10 & 0.759 & 36.92 & 0.503 \\
                         & SST & 66.29 & 0.801 & 42.57 & 0.572 \\\hline
    \multirow{2}{*}{SOD} & iGAN & 61.59 & 0.778 & 42.89 & 0.544 \\
                         & LGVT & 74.94 & 0.873 & 40.71 & 0.593 \\
    \hline
    \multirow{4}{*}{AVS} & AVS~(ResNet50) & 72.79 & 0.848 & 47.88 & 0.578 \\
                         & AVS~(PVT)     & 78.74 & 0.879 & 54.00 & 0.645 \\
                         & Ours~(ResNet50)                          & 74.13 & 0.854 & 44.95 & 0.568 \\
                         & Ours~(PVT)                               & \textbf{81.71} & \textbf{0.904} & \textbf{55.10} & \textbf{0.668} \\
\bottomrule
\end{tabular}
\vspace{-2mm}
\end{table}

\paragraph{Qualitative comparisons.} 
We offer qualitative examples to compare our framework with previous AVSBench methods~\cite{zhou2022audio}. 
The quantitative results of the S4 setting and the MS3 setting are displayed in Figure~\ref{fig:6}. 
Our proposed model accurately segments all sounding objects and outlines their shapes with precision.

\subsection{Ablation Studies}
\paragraph{Impact of the audio signal.} As illustrated in Figure~\ref{fig:3}, the audio-visual fused module is used for the audio-visual interactions from a temporal and pixel-wise level, introducing the audio information to explore the visual segmentation. We conduct an ablation study to explore its impact as shown in Table~\ref{tab: table2}. We remove the audio part of the model and disable the audio-visual fused module to explore the importance of the audio information, leading to a simple unimodal framework with only video input, the results are shown in the first row of the table. For comparison, the second row shows the proposed method with the audio-visual fused module but without the motion correlation and bidirectional generation. It is noticed that adding the audio features to the visual one leads to a significant gain under the S4 and MS3 setting. This indicates directly the audio is especially beneficial to samples with single and multiple sound sources due to the audio signals can guide which object(s) to segment. Furthermore, our proposed audio-visual fused module can also help enhance the performance over various settings.

\begin{table}[h]
\small
\centering
\caption{Impact of the audio signal. Two rows back show the proposed method with or without the audio-visual fused module.}
 \setlength{\tabcolsep}{1.2mm}
\label{tab: table2}
\vspace{-2mm}
\begin{tabular}{cccccccc}
\toprule
    \multirow{2}{*}{Variants} &\multicolumn{2}{c}{S4} &\multicolumn{2}{c}{\makecell[c]{MS3 \\From Scratch}} &\multicolumn{2}{c}{\makecell[c]{MS3 Pretrained \\on Single-Source}} \\
    \cmidrule(r){2-3} \cmidrule(r){4-5} \cmidrule(r){6-7}
     &mIoU &F-score &mIoU &F-score &mIoU &F-score \\
        \midrule
            w/o audio &80.16 &0.891 &51.85 &0.64 &55.63 &0.672 \\
         w audio &\textbf{80.82} &\textbf{0.895} &\textbf{54.35}  &\textbf{0.661} &\textbf{56.51} & \textbf{0.682}\\
\bottomrule
\end{tabular}
\end{table}

\begin{table}[!htp]
\small
\centering
\caption{Effectiveness of motion correlation.}
\label{tab: table4}
\vspace{-2mm}
 \setlength{\tabcolsep}{1.0mm}
\begin{tabular}{cccccccc}
\toprule
    \multirow{2}{*}{Variants} &\multicolumn{2}{c}{S4} &\multicolumn{2}{c}{\makecell[c]{MS3 \\From Scratch}} &\multicolumn{2}{c}{\makecell[c]{MS3 Pretrained \\on Single-Source}} \\
    \cmidrule(r){2-3} \cmidrule(r){4-5} \cmidrule(r){6-7}
     &mIoU &F-score &mIoU &F-score &mIoU &F-score \\
        \midrule									
            w/o motion & 80.82 & 0.895 & 54.35 & 0.661 & 56.51 & 0.672 \\
            with motion & \textbf{81.26} & \textbf{0.898} & \textbf{54.91} & \textbf{0.670} & \textbf{58.33} & \textbf{0.697} \\
\bottomrule
\end{tabular}
\vspace{-5mm}
\end{table}

\begin{table*}[t]\small
\centering
\caption{Effectiveness of bidirectional generation. We shorten bidirectional generation to BG. We conducted a comparison to see how much adding BG improved accuracy with and without our implicit motion estimation module. Our results show that BG greatly enhances accuracy when the motion estimation module is not present. However, when we combine motion and BG, our model performs even better. We also present the results of the AVSBench method with BG. These results demonstrate that our bidirectional generation approach is highly effective, regardless of the network structure.}
\label{tab: table5}
 \setlength{\tabcolsep}{5.8mm}
\begin{tabular}{cccccccc}
\toprule
    \multicolumn{2}{c}{Methods} &\multicolumn{2}{c}{S4} &\multicolumn{2}{c}{\makecell[c]{MS3 \\From Scratch}} &\multicolumn{2}{c}{\makecell[c]{MS3 Pretrained \\on Single-Source}} \\
    \cmidrule(r){3-4} \cmidrule(r){5-6} \cmidrule(r){7-8}
     & &mIoU &F-score &mIoU &F-score &mIoU &F-score \\
        \midrule
        \multirow{2}{*}{AVSBench} &w/o BG & 78.74 & 0.879 & 54.00 & 0.645 & 57.34 & - \\
             & with BG & \textbf{80.64} & \textbf{0.895} & \textbf{54.78} & \textbf{0.657} & \textbf{59.04} & \textbf{0.680} \\\hline
        \multirow{2}{*}{Ours w/o motion} & w/o BG & 80.82 & 0.895 & \textbf{54.35} & 0.661 & 56.51 & 0.672 \\
            & with BG & \textbf{81.85} & \textbf{0.903} & 53.49  & \textbf{0.663} & \textbf{58.81} & \textbf{0.708} \\
        \hdashline
        \multirow{2}{*}{Ours with motion} & w/o BG & 81.26 & 0.898 & 54.91 & \textbf{0.670} & 58.33 & 0.697 \\
            & with BG & \textbf{81.71} & \textbf{0.904} & \textbf{55.10} & 0.668 & \textbf{58.58} & \textbf{0.715} \\
\bottomrule
\end{tabular}
\end{table*}

\begin{figure*}
\begin{center}
\centering
  \includegraphics[width=1\linewidth]{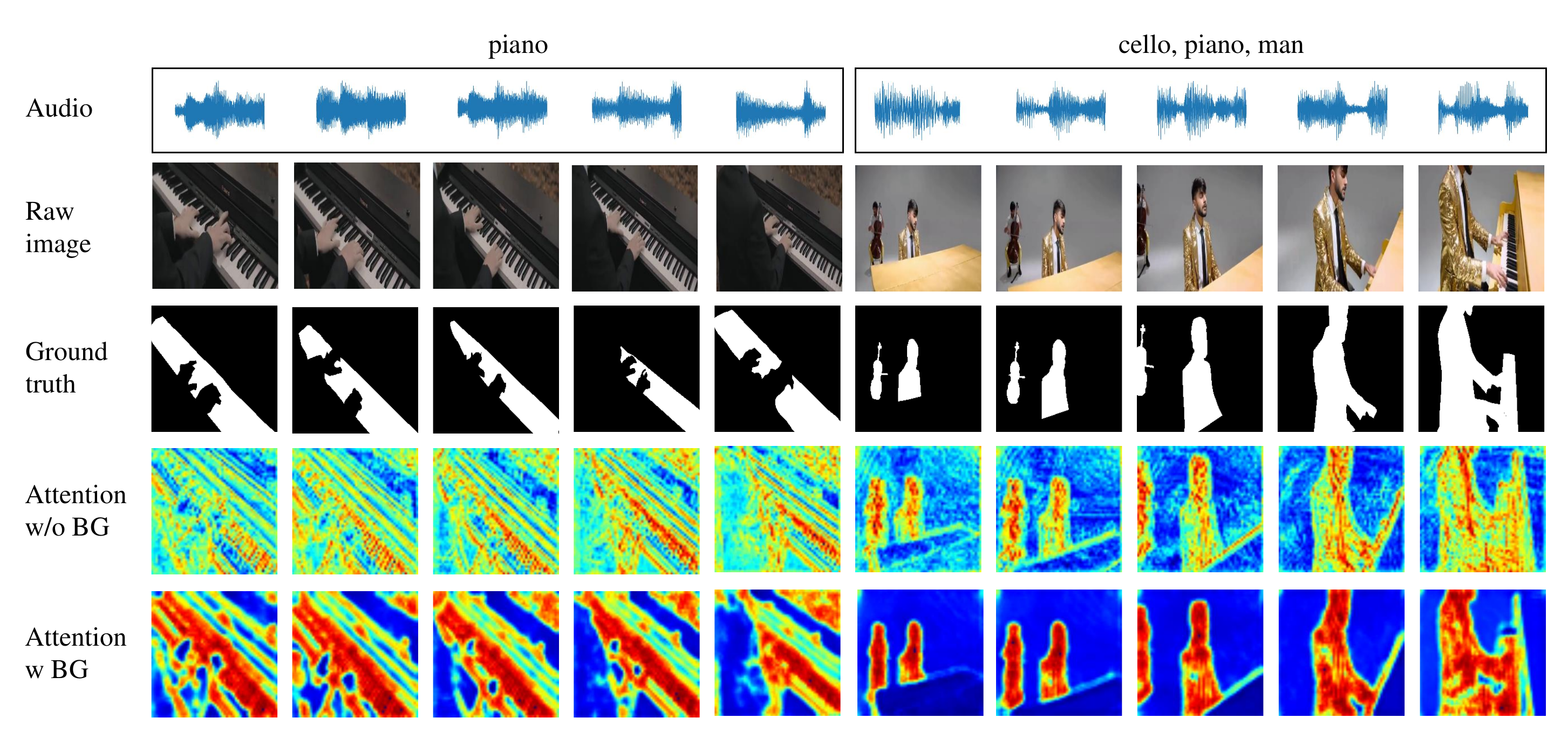}
  \vspace{-6mm}
  \caption{Audio-visual attention maps from motion correlation module under the multi-sources setting. A brighter color indicates a higher response. This indicates that our bidirectional generation constraint assists the model in focusing more on the visual regions that correspond to the audio. }
  \vspace{-2mm}
  \label{fig:8}
  \end{center}
\end{figure*}

\begin{table*}[!htp]\small
\centering
\caption{Comparison with methods from different backbones and different initialization strategies under the MS3 settings. We report the performance with ResNet50 and PVT-v2 as a backbone for the results of AVSBench and Ours under S4 and MS3 settings. We also report performance with different initialization strategies under the MS3 setting.}
\label{tab: table7}
 \setlength{\tabcolsep}{6mm}
\begin{tabular}{ccccccc}
\toprule
    \multirow{2}{*}{Metrics} & \multirow{2}{*}{Setting} & \multicolumn{2}{c}{AVSBench} & \multicolumn{2}{c}{Ours}\\
    \cmidrule(r){3-4} \cmidrule(r){5-6}
    & & ResNet50 & PVT-v2 & ResNet50 & PVT-v2 \\
    \midrule
    \multirow{3}{*}{mIoU} & S4 & 72.79 & 78.74 & 74.13 & \textbf{81.71} \\
                          & \makecell[c]{MS3 From Scratch} &47.88	&54.00& 44.95 & \textbf{55.10} \\
                          & \makecell[c]{MS3 Pretrained on Single-Source}&54.33 & 57.34 & 49.58 & \textbf{58.58} \\
    \hdashline
    \multirow{3}{*}{F-score} & S4 & 0.848 &	0.879 &0.854 & \textbf{0.904} \\
                             & \makecell[c]{MS3 From Scratch} & 0.578 & 0.645 & 0.568 & \textbf{0.668} \\
                             & \makecell[c]{MS3 Pretrained on Single-Source} & - & - & 0.627 & \textbf{0.715} \\
\bottomrule
\end{tabular}
\end{table*}

\paragraph{Effectiveness of motion correlation.}~We expect that the visual correlation module will estimate the movement information of intra-frames to enhance the network's accuracy to segment the correct objects. Therefore, we propose a motion correlation module, as shown in Figure~\ref{fig:4}. As shown in Table~\ref{tab: table4}, smoothing big-angle movement achieves a clear performance gain. For example, Our method with motion correlation but without bidirectional generation improves the mIoU and F-score slightly. This demonstrates the benefits of introducing such a motion correlation.

\paragraph{Effectiveness of bidirectional generation.} We conduct experiments to ablate the influence of cycle consistency constraints, as shown in Table~\ref{tab: table5}. Under the dotted line, we reserve the motion estimation branch and explore the effectiveness of bidirectional generation. Although in the network framework with only a motion correlation module, our model already has a good segmentation performance. After introducing bidirectional generation, our model further improves the mIoU by around $0.25$ and the F-score by about $0.018$ in the MS3 setting pre-trained on the S4. Disabling motion estimation leads to even greater improvement when using bidirectional generation constraints.

To further verify the effectiveness of our proposed cycle-consistency constraint, we added bidirectional generation to the original AVSBench, and the experimental results are shown in Table~\ref{tab: table5}. It is observed that the performance of the model can be significantly improved by using our proposed cycle consistency constraint while keeping the original AVSBench framework unchanged.

Besides, we also visualize the audio-visual attention matrices to explore what happens in the cycle-consistency constraint process. In detail, the attention map is obtained from the visual correlation attention module. We upsample it to have the same shape as the video frame. As shown in Figure~\ref{fig:8}, the high response area basically overlaps the region of sounding objects. It suggests that our bidirectional generation constraint builds a mapping from the visual pixels to the audio signals, which is semantically consistent.

\subsection{Discussion of the model training and inference}
\paragraph{Comparison with methods from different backbones.}~We compare the performance under different backbones with ResNet50 and PVT-v2. The performance on mIoU and F-score 
is reported in Table~\ref{tab: table1}. It indicates that our framework has a significantly superior segmentation performance than the previous method. The performance gain comes from our designed visual correlation attention module and bidirectional generation constraint. 

\paragraph{Pre-training on the Single-source Subset.}~
Motivated by AVSBench~\cite{zhou2022audio}. The AVSBench model at the MS3 setting has been enhanced from $54.00$ to $55.10$. We conducted experiments by initializing model parameters through pre-training on S4 dataset. The positive impact becomes more evident as the mIoU increases from $55.10$ to $58.58$ and the F-score increases from $0.668$ to $0.715$. Additionally, it has been proven that the pre-training strategy is advantageous in all settings.
\begin{table}[!htp]
\small
\centering
\caption{Parameters and inference time. Our method achieves better accuracy while requiring fewer parameters and with a faster inference time.}
\vspace{-2mm}
\label{tab: table8}
 \setlength{\tabcolsep}{5.8mm}
\begin{tabular}{ccc}
\toprule
    Methods & Parameters~(M) & Time~(ms)\\
        \midrule
            AVSBench & 101.32 & 43.55 \\
            Ours                          & 85.50  & 27.48 \\
\bottomrule
\end{tabular}
\vspace{-2mm}
\end{table}

\paragraph{Parameters and inference time.}~We have included a comparison of our parameters and inference time with AVSBench~\cite{zhou2022audio} in Table~\ref{tab: table8}. Our audio-visual fused module has replaced the TPAVI~\cite{zhou2022audio} module, and we have reduced the number of neck channels from 256 to 128. This has resulted in a decrease in parameter numbers and a faster inference speed.

\section{Conclusion}
This paper presented a novel approach to audio-visual segmentation (AVS) that addresses the limitations of traditional methods by leveraging a bidirectional generation framework. 
By capitalizing on the human ability to mentally simulate the relationship between visual characteristics and associated sounds, our approach establishes robust correlations between these modalities, leading to enhanced AVS performance. 
The introduction of a visual-to-audio projection component, capable of reconstructing audio features from object segmentation masks, showcases the effectiveness of our methodology in capturing intricate audio-visual relationships. Additionally, the implicit volumetric motion estimation module tackles the challenge of temporal dynamics, particularly relevant for sound-object motion connections. 
Through comprehensive experiments and analyses conducted on the AVSBench benchmark, we demonstrated our approach's superiority, achieving a new state-of-the-art performance level, particularly excelling in complex scenarios involving multiple sound sources. 
Our work paves the way for further advancements in AVS. 
Future research could focus on refining the bidirectional generation framework, exploring novel ways to capture nuanced audio-visual associations, and investigating applications beyond AVS. 

\section{Acknowledgments}
This research was supported in part by the National Natural Science Foundation of China (62271410), the National Key R\&D Program of China (NO.2022ZD0160100), and the Fundamental Research Funds for the Central Universities. 

\bibliography{aaai24}

\end{document}